\documentclass[letterpaper,10pt,journal,twoside]{IEEEtran}

\IEEEoverridecommandlockouts

\usepackage{graphicx}
\usepackage{amsmath,amssymb}
\usepackage{array}
\usepackage{multirow}
\usepackage{cite}
\usepackage{microtype}
\usepackage[caption=false,font=normalsize,labelfont=sf,textfont=sf]{subfig}
\usepackage[hidelinks]{hyperref}
\usepackage{orcidlink}

\title{
Collision-Free Velocity Scheduling for Multi-Agent Systems on Predefined Routes via Inexact-Projection ADMM
}

\author{Seungyeop Lee\orcidlink{0009-0001-4704-7805} and Jong-Han Kim\orcidlink{0000-0002-9030-0490}, \emph{Member, IEEE}

\thanks{Received 21 March 2026; revised 14 June 2026; accepted 13 July 2026. Date of publication 00 July 2026; date of current version 00 July 2026. This article was recommended for publication by Editor Chao-Bo Yan upon evaluation of the associate editor and reviewers’ comments. 
This work was supported in part by the Unmanned Vehicles Core Technology R\&D Program through the National Research Foundation of Korea (NRF) and the Unmanned Vehicle Advanced Research Center, funded by the Ministry of Science and ICT (MSIT), South Korea (NRF-2020M3C1C1A02086425),
and in part by a grant from the Institute of Information \& Communications Technology Planning \& Evaluation (IITP), funded by MSIT, South Korea (RS-2026-25548560, Artificial Intelligence Innovation Graduate School, Inha University).
\emph{(Corresponding author: Jong-Han Kim)}.
}

\thanks{The authors are with the Department of Aerospace Engineering and the Program in Aerospace Systems Convergence, Inha University, Incheon, South Korea. \texttt{seungyeoplee@inha.edu;\,jonghank@inha.ac.kr}}%
\thanks{Jong-Han Kim is also with the Aerospace Systems Research Institute, Inha University, Incheon, South Korea.}%
\thanks{Source code and interactive video demonstrations provided by the authors are available at \url{https://github.com/lsy010119/MACSPO}}%

\thanks{Digital Object Identifier 10.1109/LRA.2026.0000000}
}

\DeclareMathOperator*{\minimize}{\mathrm{minimize}}
\DeclareMathOperator*{\subjectto}{\mathrm{subject}\;\mathrm{to}}
\DeclareMathOperator*{\argmin}{\mathrm{arg}\,\mathrm{min}\,}

\begin{document}

\maketitle


\begin{abstract}
In fixed-route multi-agent systems, such as urban air mobility and
warehouse logistics, collision avoidance relies on temporal scheduling
rather than spatial rerouting. Existing fixed-route coordination methods
often rely on priority assignment, explicit sequencing, or mixed-integer
order selection. However, selecting an effective passing-order can be
difficult, especially when additional waypoint-level timing requirements
such as dwell times or fixed-terminal times are present. This paper
presents a priority-free temporal scheduling framework for fixed-route
multi-agent coordination. The framework optimizes waypoint passage times
to minimize the sum of terminal arrival times while satisfying speed limits, timing
specifications, and collision avoidance constraints. To handle collision
avoidance, a smooth flyby surrogate maps discrete passage times into
continuous spatial trajectories. The resulting nonlinear and nonconvex scheduling problem is solved using an inexact-projection ADMM scheme without introducing explicit passing-order variables. Numerical comparisons with order-based and reactive baselines,
together with additional stress tests, are presented to evaluate
the performance and limitations of the proposed approach.
\end{abstract}


\section{Introduction}

\IEEEPARstart{C}{ollision} avoidance is a fundamental requirement in multi-agent systems and has been widely studied across various motion planning and control frameworks \cite{bareiss2015generalized,bareiss2013reciprocal,best2016real,borrmann2015control,ames2016control,pang2020distance,baca2018model,hu2020convergent}. However, existing methods primarily resolve conflicts by modifying spatial trajectories. In domains such as urban air mobility and warehouse logistics, agents typically operate on fixed-routes where lateral maneuvers are restricted. Consequently, spatial avoidance strategies are inherently limited in
these systems, motivating approaches that resolve conflicts by adjusting timing of agents along their predefined paths.

Fixed-route conflicts are conventionally addressed through priority assignment, conflict-based search, or discrete sequencing variables. Priority and safe-interval planning algorithms sequence agents incrementally, treating previously planned trajectories as dynamic obstacles \cite{okumura2022priority,ali2023safe}. Mixed-integer and sequential convex programming approaches formulate passing interactions via binary or logical constraints before optimizing timing or speed profiles \cite{augugliaro2012generation,kushleyev2012towards,huang2019cooperative,wu2019temporal}. Recent fixed-route timing and cooperative speed planning strategies also adjust speed profiles along prescribed paths, frequently relying on priority queues or explicit ordering heuristics \cite{mao2024topp,cosineopt2026}. While these frameworks perform well when an effective passing-order can be identified efficiently, finding such an order generally requires combinatorial search, and the resulting performance can depend strongly
on the selected order. Furthermore, integrating additional temporal constraints—such as dwell times or fixed-terminal times—substantially expands the discrete decision space \cite{zhang2022priority,ma2018mapfdl,gao2023mapftw}.

Motivated by these limitations, we propose a priority-free temporal optimization framework that directly resolves fixed-route conflicts without relying on discrete sequencing variables. By formulating multi-agent coordination strictly as a waypoint passage time optimization problem, the spatial route geometry is preserved. Under this formulation, speed limits map to inter-waypoint duration bounds, and specific temporal constraints are explicitly incorporated. To enforce pairwise collision avoidance constraints over the mission
horizon, the framework employs a differentiable continuous-time trajectory surrogate. The surrogate maps waypoint passage times to continuous position trajectories, thereby enabling first-order optimization of the sampled safety constraints. This surrogate maps waypoint passage times to continuous spatial trajectories, which enables the integration of spatial collision avoidance constraints into the optimization problem and facilitates the application of gradient-based optimization techniques.

The primary contributions of this work are summarized as follows:
\begin{enumerate}
    \item We formulate multi-agent coordination on fixed-routes as a waypoint passage time optimization problem. The formulation uses only timing variables and directly incorporates speed limits, dwell times, and optional terminal time constraints.
    \item We introduce a differentiable flyby surrogate for fixed-route motion. The surrogate maps waypoint passage times to continuous position trajectories, enabling first-order, distance-based collision avoidance corrections in the timing variables.
    \item We develop a residual-controlled inexact alternating direction
    method of multipliers (ADMM) solver with an interval closest-approach
    safety penalty and evaluate it against order-based and reactive baselines.
\end{enumerate}

We evaluate the proposed framework against representative fixed-route
baselines on standard multi-agent benchmarks, timing-constrained cases,
and random-crossing stress tests.

\section{Problem Statement}

\subsection{Notation}
Throughout this paper, scalars are denoted by nonbold symbols, vectors by lowercase bold symbols, and matrices by uppercase bold symbols. The set of integers $\{1,2,\ldots,N\}$ is denoted by $[N]$. For $\mathbf{x}\in\mathbb{R}^N$, the difference operator $\Delta$ is defined as $\Delta \mathbf{x}:=\mathbf{x}_{2:N}-\mathbf{x}_{1:N-1}$, where $\mathbf{x}_{i:j}$ denotes the subvector from the $i$-th to the $j$-th element. Vertical concatenation is denoted by $[\mathbf{x}_1;\mathbf{x}_2;\ldots;\mathbf{x}_N]:=[\mathbf{x}_1^T,\mathbf{x}_2^T,\ldots,\mathbf{x}_N^T]^T$. For a scalar $a$, let $[a]_+:=\max(a,0)$. The indicator function of a set $\mathcal{C}$ is denoted by $I_{\mathcal{C}}(\mathbf{x})$, where $I_{\mathcal{C}}(\mathbf{x})=0$ if $\mathbf{x}\in\mathcal{C}$ and $I_{\mathcal{C}}(\mathbf{x})=+\infty$ otherwise.

Table~\ref{tab:notations} summarizes the main variables and parameters. 

\begin{table}[b]
\caption{Notation}
\label{tab:notations}
\centering
\renewcommand{\arraystretch}{1.15}
\begin{tabular}{ll}
\hline
\textbf{Variable} & \textbf{Description} \\
\hline
$K$ & Number of agents \\
$N_i$ & Number of waypoints assigned to agent $i$ \\
$N$ & Number of timing variables, $N=\sum_{i=1}^{K}N_i$ \\
$\mathbf{p}^{(i)}_n$ & $n$-th spatial waypoint for agent $i$ \\
$t^{(i)}_n$ & Passage time of agent $i$ at $\mathbf{p}^{(i)}_n$ \\
$\mathbf{t}^{(i)}$ & Passage time vector of agent $i$, $[t^{(i)}_1,\ldots,t^{(i)}_{N_i}]^T$ \\
$\Delta t^{(i)}_n$ & Duration of segment $n$ for agent $i$ \\
$d^{(i)}_n$ & Length of segment $n$ for agent $i$ \\
$\mathbf{v}^{(i)}_n$ & Nominal velocity of agent $i$ on segment $n$ \\
$v_{\rm min}, v_{\rm max}\!\!$ & Prescribed speed bounds \\
$t_{\rm s}^{(i)}$ & Prescribed departure time of agent $i$ \\
$t_{\rm f}^{(i)}$ & Prescribed terminal time of agent $i$, if imposed \\
$\mathcal{K}_{\rm f}$ & Set of agents with fixed-terminal times \\
$\mathcal{K}_{\rm u}$ & Set of agents with free terminal times, $[K]\setminus\mathcal{K}_{\rm f}$ \\
$\mathcal{P}$ & Set of unordered agent pairs \\
$T_{\rm max}$ & Maximum terminal arrival time, $T_{\rm max}\!:=\!\max_{i\in[K]} t^{(i)}_{N_i}\!$ \\
$\mathcal{T}$ & continuous-time domain, $\mathcal{T}:=[0,T_{\rm max}]$ \\
$\mathcal{T}_{\rm d}$ & Uniform discrete temporal grid of $\mathcal{T}$ \\
$M$ & Number of discrete time nodes, $M:=|\mathcal{T}_{\rm d}|$ \\
\hline
\end{tabular}
\end{table}

\subsection{Assumptions}
We consider a system of $K$ agents. Each agent $i$ is assigned a
predefined waypoint sequence $\{\mathbf p_n^{(i)}\}_{n=1}^{N_i}$ and follows the corresponding piecewise-linear path using flyby maneuvers.

Each agent is assumed to move at a constant speed along each segment. Under this assumption, the velocity on segment $n$ is parameterized by the waypoint passage times as follows:
\begin{equation*}
\mathbf{v}^{(i)}_n(\mathbf{t}^{(i)}) =
\frac{\mathbf{p}^{(i)}_{n+1}-\mathbf{p}^{(i)}_{n}}{\Delta t^{(i)}_n},
\qquad n=1,\ldots,N_i-1.
\end{equation*}

\begin{figure}
    \centering
    \includegraphics[width=\linewidth]{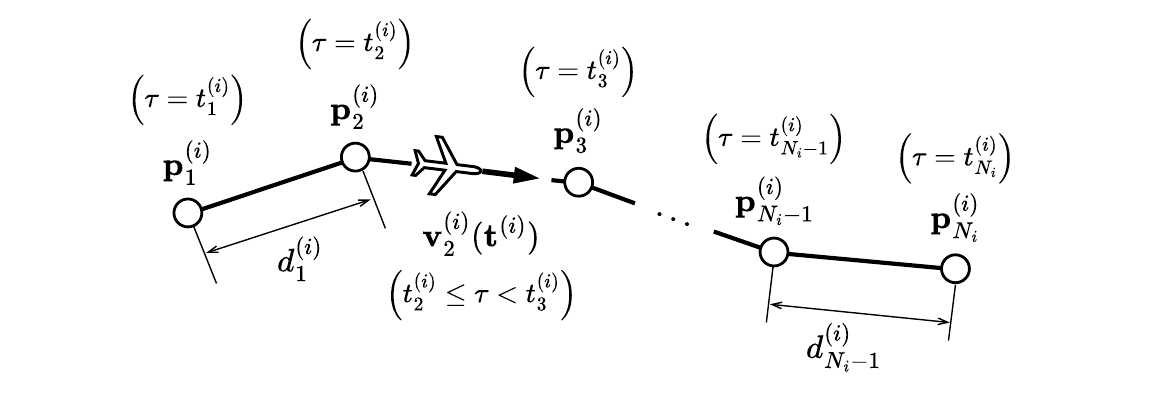}
    \caption{Illustration of a piecewise-linear waypoint path and its corresponding segment notation.}
    \label{trajexample}
\end{figure}

Fig.~\ref{trajexample} illustrates the resulting waypoint path representation.

\subsection{Problem Formulation}

We formulate a fixed-route velocity-scheduling problem in which the
waypoint passage times $\{\mathbf{t}^{(i)}\}_{i=1}^{K}$ are the decision variables. The objective is to
minimize the sum of terminal arrival times of agents with free terminal
times, subject to boundary-time constraints, speed bounds, and pairwise
safety constraints.

Each agent departs at a prescribed time $t_{\rm s}^{(i)}$. Terminal constraints apply only to agents in $\mathcal{K}_{\rm f}$, whereas agents in $\mathcal{K}_{\rm u}$ contribute to the arrival time objective. Under the constant speed assumption, speed limits are expressed as box constraints on the segment durations. Finally, collision avoidance is imposed on the continuous spatial trajectories for every agent pair in $\mathcal{P}$ at every temporal-grid point in $\mathcal T_{\rm d}$.

These conditions give
\begin{equation}\label{prob1}
\begin{aligned}
& \minimize_{\{\mathbf{t}^{(i)}\}_{i=1}^{K}} && \sum_{i\in \mathcal{K}_{\rm u}} t^{(i)}_{N_i} \\
& \subjectto && t^{(i)}_1 = t^{(i)}_{\rm s}, \quad \forall i\in[K],\\
& && t^{(i)}_{N_i} = t^{(i)}_{\rm f}, \quad \forall i\in \mathcal{K}_{\rm f},\\
& && \frac{\mathbf{d}^{(i)}}{v_{\rm max}} \le \Delta \mathbf{t}^{(i)} \le \frac{\mathbf{d}^{(i)}}{v_{\rm min}}, \quad \forall i\in[K],\\
& && \big\|\mathbf{p}^{(i)}(\mathbf{t}^{(i)},\tau)-\mathbf{p}^{(j)}(\mathbf{t}^{(j)},\tau)\big\|_2 \ge d_{\rm safe},\\
& && \hspace{22mm} \forall (i,j)\in\mathcal{P},\ \forall \tau\in\mathcal{T}_{\rm d}.
\end{aligned}
\end{equation}
Here, $\Delta \mathbf{t}^{(i)} := [\Delta t^{(i)}_1, \ldots, \Delta t^{(i)}_{N_i-1}]^T$ and $\mathbf{d}^{(i)} := [d^{(i)}_1, \ldots, d^{(i)}_{N_i-1}]^T$ denote the segment duration and length vectors, respectively. The vector $\mathbf{p}^{(i)}(\mathbf{t}^{(i)},\tau)$ denotes the position of agent $i$ at time $\tau$ induced by its waypoint sequence and passage times $\mathbf{t}^{(i)}$.

\section{Surrogate Trajectory Model}

A piecewise-linear trajectory exhibits instantaneous velocity changes
at waypoints, producing a nonsmooth inter-agent distance function. We
therefore replace abrupt segment switching with a smooth surrogate.

Given the prescribed waypoints and passage times, the surrogate velocity of agent $i$ is
\begin{equation}\label{vapprox}
\tilde{\mathbf{v}}^{(i)}(\mathbf{t}^{(i)},\tau)
\!=\!\!\!
\sum_{n=1}^{N_i-1}\!\!
\mathbf{v}^{(i)}_n(\mathbf{t}^{(i)})
\Big[
\sigma\big(\beta(\tau-\hat t^{(i)}_n)\big)
-
\sigma\big(\beta(\tau-\hat t^{(i)}_{n+1})\big)
\Big],
\end{equation}
where $\sigma(x)=1/(1+e^{-x})$ is the sigmoid function, $\hat{t}^{(i)}_n := t^{(i)}_n + 2\ln(2)/\beta$ is the shifted transition center, and $\beta$ is the transition sharpness parameter. The constant offset centers the transition in time.

Integrating \eqref{vapprox} gives the surrogate position model
\begin{multline}\label{papprox}
\tilde{\mathbf{p}}^{(i)}(\mathbf{t}^{(i)},\tau)
=
\mathbf{p}^{(i)}_1
\!\!+\!\!
\sum_{n=1}^{N_i-1}\!
\frac{\mathbf{v}^{(i)}_n(\mathbf{t}^{(i)})}{\beta}
\Big[
\zeta\big(\beta(\tau-\hat t^{(i)}_n)\big)
\\[-4pt]
-\zeta\big(\beta(\tau-\hat t^{(i)}_{n+1})\big) - b_n^{(i)}
\Big],
\end{multline}
where $\zeta(x):=\ln(1+e^x)$ is the softplus function,
and $b_n^{(i)} := \zeta\big(\beta(t^{(i)}_1-\hat t^{(i)}_n)\big)
-\zeta\big(\beta(t^{(i)}_1-\hat t^{(i)}_{n+1})\big)$. 
The trajectory in \eqref{papprox} is differentiable with respect to $\tau$ and $\mathbf{t}^{(i)}$, so the collision constraint in \eqref{prob1} can be handled using first-order information.

In many operational scenarios, agents are required to remain stationary at specific waypoints for designated dwell times. These dwell-time requirements are incorporated by shifting the segment-transition intervals in the surrogate model, rather than by introducing additional ADMM constraints. This parameterization encodes the prescribed dwell durations while preserving the split structure of the optimization problem.

Let $t_{{\rm wp},n}^{(i)}\ge0$ denote the dwell time at waypoint $n$ before departing along segment $n$, with $t_{{\rm wp},1}^{(i)}=0$, and define
\begin{equation*}
\kappa_n^{(i)}
:=
\sum_{\ell=1}^{n} t_{{\rm wp},\ell}^{(i)},
\qquad n=1,\ldots,N_i-1.    
\end{equation*}

In the surrogate evaluation, the transition interval for segment $n$ is shifted from $[t_n^{(i)},t_{n+1}^{(i)}]$ to
\begin{equation*}
\left[
t_n^{(i)}+\kappa_n^{(i)},\;
t_{n+1}^{(i)}+\kappa_n^{(i)}
\right].    
\end{equation*}

Equivalently, the two transition centers in the $n$-th term of \eqref{vapprox} and \eqref{papprox} are replaced by
\begin{equation*}
\hat t_{n,{\rm L}}^{(i)}
=
t_n^{(i)}
+
\kappa_n^{(i)}
+
\frac{2\ln2}{\beta},
\quad
\hat t_{n,{\rm R}}^{(i)}
=
t_{n+1}^{(i)}
+
\kappa_n^{(i)}
+
\frac{2\ln2}{\beta}.    
\end{equation*}

This convention inserts a prescribed dwell interval between consecutive
moving segments while retaining the waypoint timing variables $\mathbf t^{(i)}$. The executed terminal time of agent $i$ is $t_{N_i}^{(i)}+\kappa_{N_i-1}^{(i)}$. Therefore, a prescribed terminal
time $t_{\rm f}^{(i)}$ is imposed through
\begin{equation*}
t_{N_i}^{(i)}
+
\kappa_{N_i-1}^{(i)}
=
t_{\rm f}^{(i)}.    
\end{equation*}

The surrogate model \eqref{papprox} generates a flyby path near the waypoints, and the miss distance is controlled by $\beta$. To keep the tracking error below a prescribed tolerance $\epsilon_{\rm wp}$, $\beta$ must be chosen accordingly.

Let $\Delta \mathbf{v}^{(i)}_k := \mathbf{v}^{(i)}_k-\mathbf{v}^{(i)}_{k-1}$ denote the velocity jump at waypoint $k$, and let $\rho_\beta(s):=\zeta(-\beta |s|)/\beta$ denote the integral tail of the sigmoid transition at time offset $s$. The spatial deviation at the transition center $\hat{t}^{(i)}_k$ satisfies
\begin{equation*}
\|
\tilde{\mathbf{p}}^{(i)}(\mathbf{t}^{(i)},\hat t^{(i)}_k)
-
\mathbf{p}^{(i)}_k
\|_2
\le
\sum_{\ell=2}^{N_i-1}
\|
\Delta \mathbf{v}^{(i)}_{\ell}
\|_2\,
\rho_\beta
\!\big(
\hat t^{(i)}_k-\hat t^{(i)}_\ell
\big).    
\end{equation*}

The term with $\ell=k$ gives the dominant local miss distance.

For a given timing profile, the following condition enforces the deviation bound $\epsilon_{\rm wp}$:
\begin{equation*}
\max_{i\in[K]}\max_{k=2,\ldots,N_i-1}
\sum_{\ell=2}^{N_i-1}
\|
\Delta \mathbf{v}^{(i)}_{\ell}
\|_2\,
\rho_\beta
\!\big(
\hat t^{(i)}_k-\hat t^{(i)}_\ell
\big)
\le
\epsilon_{\rm wp}.
\end{equation*}

The exact velocity jumps depend on the timing variables. We therefore select $\beta$ using the following geometry-based local rule. Let $\theta^{(i)}_k$ denote the angle between consecutive route segments at waypoint $k$. The maximum possible velocity jump is bounded by
\begin{multline*}
D(\theta)
=
\max\Big\{
\sqrt{
v_{\rm max}^2+v_{\rm min}^2
-
2v_{\rm max}v_{\rm min}\cos\theta
}\, ,
\\[-4pt]
2v_{\rm max}\sin\frac{\theta}{2}
\Big\}.
\end{multline*}

\begin{figure}[!t]
\centering
{\includegraphics[width=1.0\linewidth]{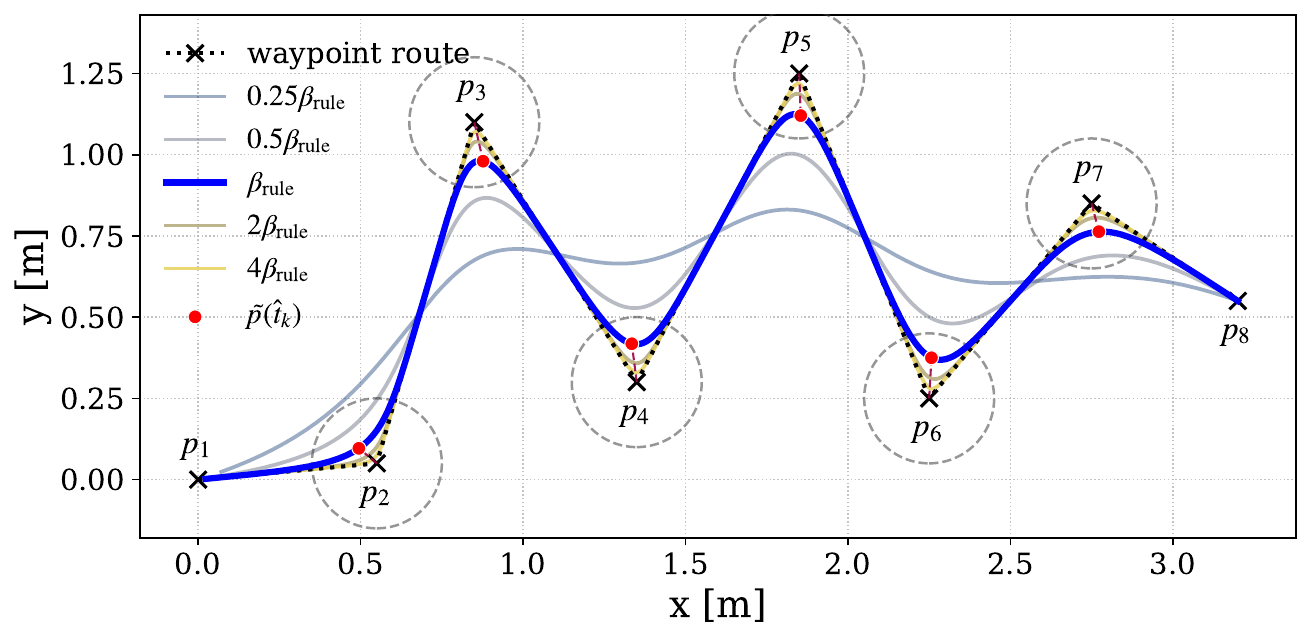}}
\caption{
    Smooth flyby surrogate in \eqref{papprox}. The blue curve corresponds to $\beta=\beta_{\mathrm{rule}}$, where $\beta_{\mathrm{rule}}$ is defined in \eqref{eq:beta_rule}. Dashed circles indicate the waypoint deviation tolerance $\varepsilon_{\mathrm{wp}}$. In this example, the surrogate trajectories with $\beta\ge\beta_{\mathrm{rule}}$ remain within the prescribed
tolerance.}
\label{fig:surrogate}
\end{figure}

To obtain a simple local rule, we neglect overlap between neighboring
transitions and retain only the largest local velocity jump. This gives
\begin{equation}\label{eq:beta_rule}
    \beta
    \ge
    \frac{\ln 2}{\epsilon_{\rm wp}}
    \max_{i,k}D(\theta^{(i)}_k).
\end{equation}
In the numerical examples, $\beta$ is fixed before optimization using \eqref{eq:beta_rule} for a given tolerance $\epsilon_{\rm wp}$. Fig.~\ref{fig:surrogate} shows an example.

\section{First-Order Optimization}

All constraints in \eqref{prob1} except collision avoidance are affine or box constraints. We use an ADMM framework to separate the timing and speed-bound terms from the collision avoidance terms~\cite{boyd2011distributed}.

\subsection{Split Form}

Define the stacked timing vector
\begin{equation*}
\mathbf{t} := \big[\mathbf{t}^{(1)}; \ldots; \mathbf{t}^{(K)}\big] \in \mathbb{R}^N.
\end{equation*}

We introduce $\mathbf x$ as the stacked segment duration vector and
$\mathbf z$ as a local copy of $\mathbf t$ used in the collision avoidance terms. Let $\mathbf{D}$ denote the stacked forward difference operator, so that
\begin{equation*}
\mathbf{D}\mathbf{t} = \mathbf{x} = \big[\Delta \mathbf{t}^{(1)}; \ldots; \Delta \mathbf{t}^{(K)}\big].
\end{equation*}

A selector matrix $\mathbf{E}$ imposes the prescribed departure time constraints and the fixed-terminal time constraints:
\begin{equation*}
\mathbf{E}\mathbf{t} = \mathbf{e}
= \big[t^{(1)}_{\rm s}, \ldots, t^{(K)}_{\rm s}, \{t^{(i)}_{\rm f}\}_{i \in \mathcal{K}_{\rm f}}\big]^T.
\end{equation*}

The objective is represented by $\mathbf{q} \in \mathbb{R}^N$ satisfying
\begin{equation*}
\mathbf{q}^T\mathbf{t} = \sum_{i \in \mathcal{K}_{\rm u}} t^{(i)}_{N_i}.
\end{equation*}

The speed and safety sets are
\begin{equation*}
\begin{aligned}
\mathcal{V} &:= \big\{\mathbf{x} : \mathbf{d}/v_{\rm max} \le \mathbf{x} \le \mathbf{d}/v_{\rm min}\big\}, \\
\mathcal{C} &:= \big\{\mathbf{z} :
\|\tilde{\mathbf{p}}^{(i)}(\mathbf{z}^{(i)},\tau)
-\tilde{\mathbf{p}}^{(j)}(\mathbf{z}^{(j)},\tau)\|_2
\ge d_{\rm safe},
\\ &\qquad\qquad\qquad\qquad\qquad\qquad\quad \forall (i,j)\in\mathcal{P},\ \forall \tau\in\mathcal{T}_{\rm d}\big\}.
\end{aligned}
\end{equation*}

Using the indicator functions $I_{\mathcal{V}}$ and $I_{\mathcal{C}}$, the split problem is
\begin{equation}\label{eq:admm_form}
\begin{aligned}
&\minimize_{\mathbf{t},\mathbf{x},\mathbf{z}} 
&& \mathbf{q}^T\mathbf{t} + I_{\mathcal{V}}(\mathbf{x}) + I_{\mathcal{C}}(\mathbf{z}) \\
&\subjectto &&
\mathbf{E}\mathbf{t} = \mathbf{e}, \\
& &&\mathbf{D}\mathbf{t} - \mathbf{x} = \mathbf{0}, \\
& &&\mathbf{t} - \mathbf{z} = \mathbf{0}.
\end{aligned}
\end{equation}

\subsection{ADMM Updates}

Let $\rho>0$ be the ADMM penalty parameter. We use scaled dual variables $\mathbf{u}_E$, $\mathbf{u}_x$, and $\mathbf{u}_z$ for $\mathbf{E}\mathbf{t}=\mathbf{e}$, $\mathbf{D}\mathbf{t}=\mathbf{x}$, and $\mathbf{t}=\mathbf{z}$, respectively.

\textit{$\mathbf{t}$-update:}
Given $(\mathbf{x}^k,\mathbf{z}^k,\mathbf{u}_E^k,\mathbf{u}_x^k,\mathbf{u}_z^k)$, the $\mathbf{t}$-subproblem is quadratic. From its optimality condition, $\mathbf{t}^{k+1}$ is obtained by solving the sparse linear system
\begin{multline}\label{eq:t_update}
\left(
\mathbf{E}^T\mathbf{E}
+
\mathbf{D}^T\mathbf{D}
+
\mathbf{I}
\right)
\mathbf{t}^{k+1}
=
\mathbf{E}^T
\left(
\mathbf{e}-\mathbf{u}_E^k
\right) \\
+
\mathbf{D}^T
\left(
\mathbf{x}^k-\mathbf{u}_x^k
\right)  
+
\mathbf{z}^k-\mathbf{u}_z^k
-
\frac{1}{\rho}\mathbf{q}.
\end{multline}
The coefficient matrix is constant and can be factorized once.

\textit{$\mathbf{x}$-update:}
The $\mathbf{x}$-update is the projection onto the speed bounds:
\begin{equation}\label{eq:x_update}
\mathbf{x}^{k+1}
=
\Pi_{\mathcal{V}}
\left(
\mathbf{D}\mathbf{t}^{k+1}
+
\mathbf{u}_x^k
\right),
\end{equation}
where the projection is applied elementwise as
\begin{equation*}
\Pi_{\mathcal{V}}(\mathbf{y})
=
\min
\Big(
\frac{\mathbf{d}}{v_{\rm min}},
\;
\max
\Big(
\frac{\mathbf{d}}{v_{\rm max}},
\mathbf{y}
\Big)
\Big).
\end{equation*}

\textit{$\mathbf{z}$-update:}
Since  the collision avoidance feasible set is nonlinear and
nonconvex, an exact projection would require solving a nonlinear
subproblem at every ADMM iteration. We instead define a finite-grid
violation penalty and construct a local first-order correction.

For each pair $(i,j)\in\mathcal{P}$ and temporal grid point $\tau_m\in\mathcal{T}_{\rm d}$, define
\begin{equation*}
\begin{aligned}
\mathbf{r}_{ijm}(\mathbf{z})
&:=
\tilde{\mathbf{p}}^{(i)}(\mathbf{z}^{(i)},\tau_m)
-
\tilde{\mathbf{p}}^{(j)}(\mathbf{z}^{(j)},\tau_m),
\\
d_{ijm}(\mathbf{z})&:=\|\mathbf{r}_{ijm}(\mathbf{z})\|_2.
\end{aligned}
\end{equation*}

The nodal violation is
\begin{equation*}
c^{{\rm node}}_{ijm}(\mathbf{z})
:=
\Big[
1-
\frac{d_{ijm}(\mathbf{z})}{d_{\rm safe}}
\Big]_+.
\end{equation*}

To reduce sensitivity to the temporal grid phase, we also evaluate the closest approach within each interval $[\tau_m,\tau_{m+1}]$. Let
\begin{equation*}
\mathbf{e}_{ijm}
:=
\mathbf{r}_{ij,m+1}
-
\mathbf{r}_{ijm}.
\end{equation*}

Using linear interpolation of the relative position, the closest point parameter is approximated as
\begin{equation*}
s_{ijm}^{\star}
:=
{\rm clip}
\Big(
-
\frac{\mathbf{r}_{ijm}^T\mathbf{e}_{ijm}}
{\|\mathbf{e}_{ijm}\|_2^2},
0,
1
\Big).
\end{equation*}

The corresponding interval distance is
\begin{equation*}
d^{\star}_{ijm}(\mathbf{z})
:=
\|
\mathbf{r}_{ijm}
+
s_{ijm}^{\star}\mathbf{e}_{ijm}
\|_2.
\end{equation*}

The interval term is used only when the closest point lies inside the interval and yields a distance smaller than those at both endpoints. This avoids double counting nodal violations. Define
\begin{equation*}
d^{{\rm end}}_{ijm}
:=
\min\big\{d_{ijm},d_{ij,m+1}\big\}.
\end{equation*}
The interval activation indicator is
\begin{equation*}
\chi_{ijm}
=
\begin{cases}
1,
&
\begin{gathered}
\epsilon_s<s_{ijm}^{\star}<1-\epsilon_s \ \text{and}\ 
d^{\star}_{ijm}<d^{{\rm end}}_{ijm},
\end{gathered}
\\
0,
& \text{otherwise},
\end{cases}
\end{equation*}
where $\epsilon_s\in(0,1/2)$. The interval violation is
\begin{equation*}
c^{{\rm int}}_{ijm}(\mathbf{z})
:=
\chi_{ijm}
\Big[
1-
\frac{d^{\star}_{ijm}(\mathbf{z})}{d_{\rm safe}}
\Big]_+.
\end{equation*}

The collision penalty used in the $\mathbf{z}$-update is the finite-grid violation sum
\begin{equation}\label{eq:collision_penalty}
\begin{aligned}
f(\mathbf{z})
=
&
\sum_{(i,j)\in\mathcal{P}}
\sum_{m=1}^{M}
c^{{\rm node}}_{ijm}(\mathbf{z})
+
\sum_{(i,j)\in\mathcal{P}}
\sum_{m=1}^{M-1}
c^{{\rm int}}_{ijm}(\mathbf{z}).
\end{aligned}
\end{equation}

\begin{figure}[t]
    \centering
    {\includegraphics[width=0.9\linewidth]{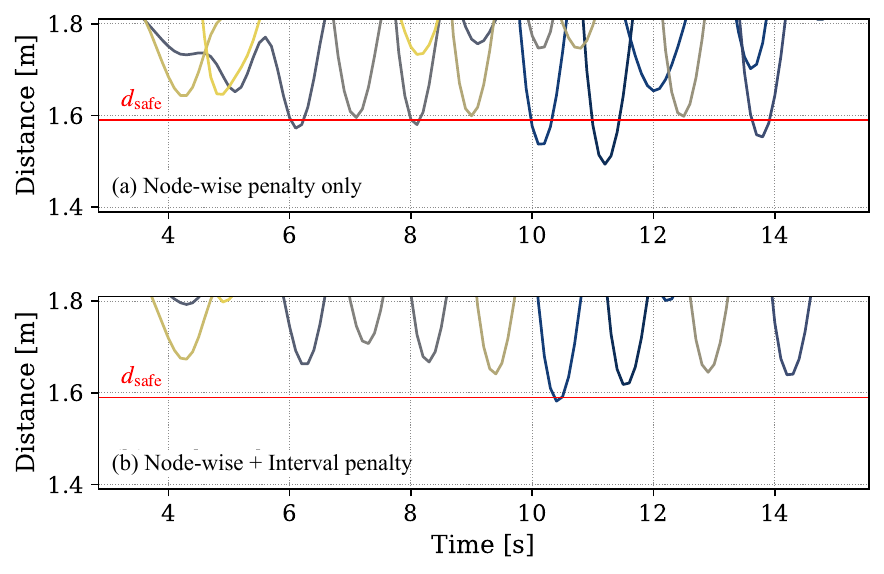}}
    \caption{
        Bottleneck example comparing the nodal penalty with the combined nodal and interval penalty. The curves show inter-agent distances for the pairs in $\mathcal P$, and the red horizontal line indicates
        $d_{\mathrm{safe}}$. In this example, the interval term eliminates the between-node violation observed when only the nodal penalty is used.
    }
    \label{fig:closest_interval_example}
\end{figure}

Fig.~\ref{fig:closest_interval_example} shows the effect of the interval closest approach term.

Replacing the collision constraints with \eqref{eq:collision_penalty} gives the following penalized proximal subproblem:
\begin{equation}\label{eq:z_exact_penalty_form}
    \mathbf{z}^{k+1}
    =
    \argmin_{\mathbf{z}}
    \;
    f(\mathbf{z})
    +
    \frac{\rho}{2}
    \|
    \mathbf{z}-
    \left(\mathbf{t}^{k+1}+\mathbf{u}_z^k\right)
    \|_2^2.
\end{equation}

The penalty $f$ is nonlinear, so solving \eqref{eq:z_exact_penalty_form} exactly at each iteration is costly. We instead compute a local correction at the proximal point $\hat{\mathbf{z}}^{k}:=\mathbf{t}^{k+1}+\mathbf{u}_z^k$. Let $f_k:=f(\hat{\mathbf{z}}^{k})$ and $\mathbf{g}_k\in\partial f(\hat{\mathbf{z}}^{k})$, where $\partial f$ denotes the subdifferential of the finite-grid hinge penalty. We linearize $f$ at $\hat{\mathbf z}^k$ as
$f(\hat{\mathbf z}^k+\Delta\mathbf z) \approx f_k+\mathbf g_k^T\Delta\mathbf z$ and define the regularized local correction by
\begin{equation*}
\Delta \mathbf{z}^k
=
\argmin_{\Delta \mathbf{z}}
\frac{1}{2}
\left(
f_k+\mathbf{g}_k^T\Delta \mathbf{z}
\right)^2
+
\frac{\delta_z}{2}
\|\Delta \mathbf{z}\|_2^2,
\end{equation*}
where $\delta_z>0$ regularizes the correction. The closed-form correction is
\begin{equation}\label{eq:z_base_correction}
\Delta \mathbf{z}^k
=
-
\frac{
f_k
}{
\|\mathbf{g}_k\|_2^2+\delta_z
}
\mathbf{g}_k,
\end{equation}
and the non-inertial update is
\begin{equation*}
\mathbf{z}^{k+1}
=
\hat{\mathbf{z}}^{k}
+
\Delta \mathbf{z}^k.
\end{equation*}

The correction in \eqref{eq:z_base_correction} reduces the linearized violation model:
\begin{equation*}
f_k+\mathbf{g}_k^T\Delta \mathbf{z}^k
=
f_k
\frac{\delta_z}{\|\mathbf{g}_k\|_2^2+\delta_z}.
\end{equation*}

Thus, for $f_k>0$ and $\mathbf{g}_k\neq\mathbf{0}$, the relaxed linearized penalty strictly decreases. However, due to the nonconvexity of $f$, this decrease property is local and does not imply monotone decrease of the original penalty. In congested scheduling scenarios, the iterates may enter a local passing-order basin in which the finite-grid violation stops decreasing sufficiently.

To reduce this stagnation, the implementation uses a gated inertial term. Let $\mathbf{m}^k$ denote the correction memory. The implemented update is
\begin{equation*}
\begin{aligned}
    \mathbf{m}^{k+1}
    &=
    \Delta \mathbf{z}^k+\alpha^k \mathbf{m}^k,\\
    \mathbf{z}^{k+1}
    &=
    \hat{\mathbf{z}}^{k}+\mathbf{m}^{k+1}.
\end{aligned}
\end{equation*}

The gain $\alpha^k$ is selected from a sliding window progress test. Let $\tilde f^k$ be the monitored penalty value, and define the lower envelope over the recent window
\begin{equation*}
    B^k
    :=
    \min_{\ell\in\{k-m_{\rm stag}+1,\ldots,k\}}
    \tilde f^\ell.
\end{equation*}

The relative decrease of the envelope is
\begin{equation*}
r_{\rm f}^k
:=
\frac{
B^{k-m_{\rm stag}}-B^k
}{
B^{k-m_{\rm stag}}
}.
\end{equation*}

The inertial term is activated only when the envelope decrease is small:
\begin{equation*}
\alpha^k=
\begin{cases}
\alpha_{\rm active},
& r_{\rm f}^k<\varepsilon_{\rm stag}\;\text{and}\;\tilde f^k>0,\\
0, & \text{otherwise}.
\end{cases}
\end{equation*}

During the inertial phase, the gain is reset to zero if the lower envelope improves or if the current penalty exceeds the envelope by a prescribed ratio:
\begin{equation*}
\alpha^k=0,
\quad
\text{if}
\quad
B^k<B^{k-m_{\rm stag}}
\ \
\text{or}
\ \
\tilde f^k>(1+\varepsilon_{\rm inc})B^k.
\end{equation*}

\textit{Dual update:}
After the three primal blocks are updated, the scaled dual variables are updated as
\begin{equation*}
\begin{aligned}
\mathbf{u}_E^{k+1}
&=
\mathbf{u}_E^k+\mathbf{E}\mathbf{t}^{k+1}-\mathbf{e},\\
\mathbf{u}_x^{k+1}
&=
\mathbf{u}_x^k+\mathbf{D}\mathbf{t}^{k+1}-\mathbf{x}^{k+1},\\
\mathbf{u}_z^{k+1}
&=
\mathbf{u}_z^k+\mathbf{t}^{k+1}-\mathbf{z}^{k+1}.
\end{aligned}
\end{equation*}

Here, $\mathbf{u}_E$, $\mathbf{u}_x$, and $\mathbf{u}_z$ correspond to the equality constraints, the speed bound, and the collision avoidance constraints, respectively.

\subsection{Convergence Remarks and Residual Floors}
The scheme is applied to a finite dimensional discretization of the continuous-time scheduling problem. The collision free set is nonconvex, and the $\mathbf{z}$-update uses an inexact gradient based correction rather than an exact projection. We therefore do not claim global optimality or convergence to a stationary point of the original continuous-time problem. The algorithm is used as a residual controlled inexact ADMM scheme.

At each iteration, we monitor the scalar primal residual
\begin{equation*}\label{eq:scalar_primal_residual}
r_{\rm pri}^{k}
:=
\max
\left\{
\|\mathbf{E}\mathbf{t}^k-\mathbf{e}\|_\infty,\,
\|\mathbf{D}\mathbf{t}^k-\mathbf{x}^k\|_\infty,\,
\|\mathbf{t}^k-\mathbf{z}^k\|_\infty
\right\},
\end{equation*}
and the maximum safety violation
\begin{equation*}\label{eq:max_safe_violation}
    r_{\rm safe}^k
    :=
    \max_{(i,j)\in\mathcal{P},\ \tau_m\in\mathcal{T}_{\rm d}}
    \Big[
    1-
    \frac{
    \|
    \mathbf{r}_{ijm}(\mathbf{z}^k)
    \|_2
    }{
    d_{\rm safe}
    }
    \Big]_+.
\end{equation*}

The iteration is terminated when
\begin{equation*}
r_{\rm pri}^{k}\le \varepsilon_{\rm pri},
\qquad
r_{\rm safe}^{k}\le \varepsilon_{\rm safe},
\end{equation*}
where $\varepsilon_{\rm pri}$ and $\varepsilon_{\rm safe}$ are prescribed residual floors. The returned iterate is accepted as an $(\varepsilon_{\rm pri},\varepsilon_{\rm safe})$ feasible solution of the finite-grid surrogate problem. The primal residual condition enforces consistency between the timing variable and the auxiliary variable, while $r_{\rm safe}^{k}$ checks the minimum pairwise distance of the trajectories.

All reported trajectories are additionally evaluated on a finer temporal grid after optimization. This process checks for constraint violations within the intervals defined by the optimization grid. In the limiting case where the residual floors are driven to zero and a convergent subsequence exists, the corresponding limit point is feasible for the finite-grid surrogate. Stationarity or local convergence of the implemented inexact $\mathbf{z}$-correction, especially with the gated inertial safeguard, would require additional regularity and subproblem accuracy assumptions.

\section{Numerical Examples}
\label{sec:numerical_examples}

\begin{figure*}[t]
    \centering
    \begin{minipage}[t]{0.245\textwidth}
        \centering
        {\includegraphics[width=\linewidth]{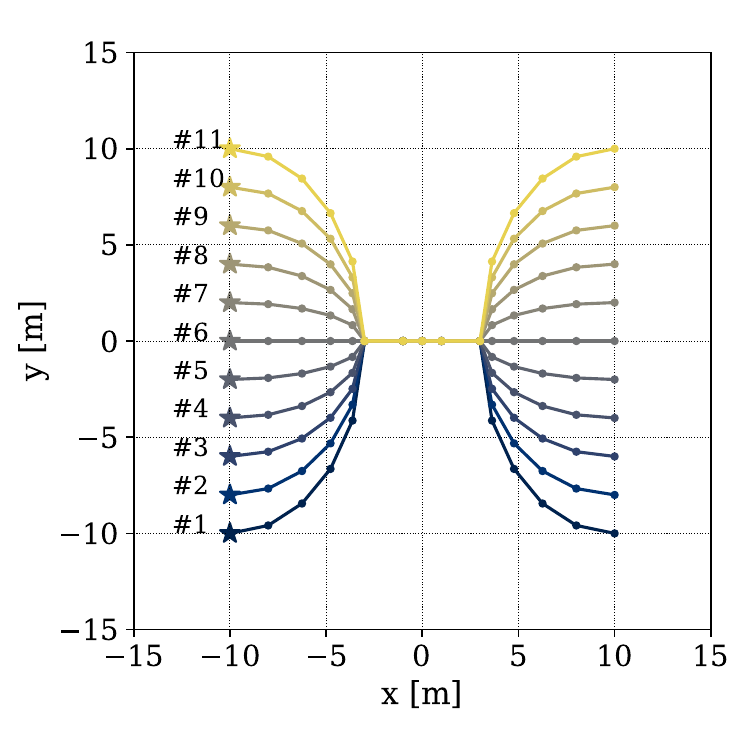}}
        \\[-8pt] {\centering (a) Bottleneck}
    \end{minipage}
    \begin{minipage}[t]{0.245\textwidth}
        \centering
        {\includegraphics[width=\linewidth]{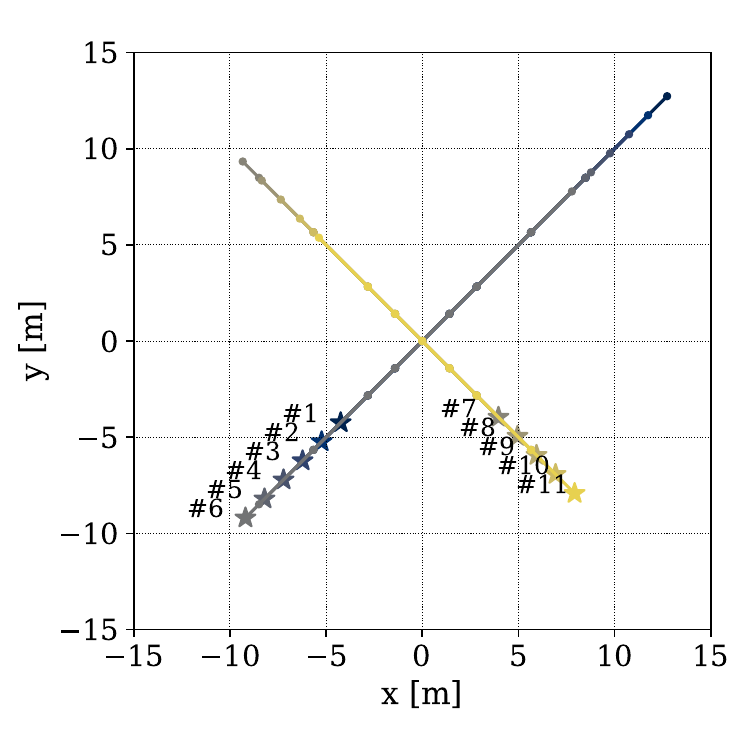}}
        \\[-8pt] {\centering (b) Aligned crossing}
    \end{minipage}
    \begin{minipage}[t]{0.245\textwidth}
        \centering
        {\includegraphics[width=\linewidth]{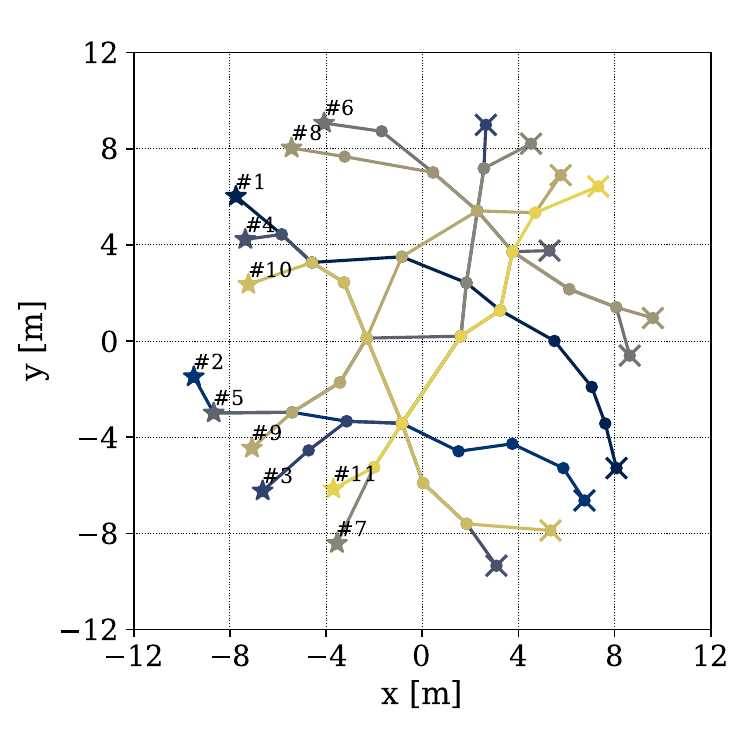}}
        \\[-8pt] {\centering (c) Graph trajectory}
    \end{minipage}
    \begin{minipage}[t]{0.245\textwidth}
        \centering
        {\includegraphics[width=\linewidth]{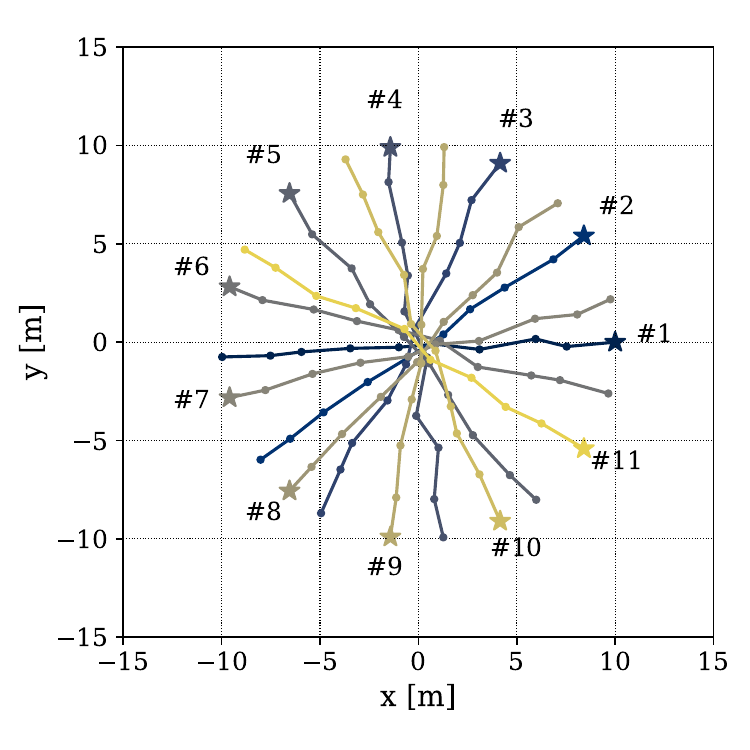}}
        \\[-8pt] {\centering (d) Random crossing}
    \end{minipage}\vspace{0.5em}
    \caption{
    Representative fixed-route benchmark geometries. The bottleneck and aligned crossing cases are deterministic. The graph trajectory and random crossing cases show representative Monte Carlo instances. Solid lines with markers denote prescribed routes.
    }
    \label{fig:benchmark_snapshots}
\end{figure*}

The experiments evaluate three aspects: performance relative to fixed-route
baselines in structured environments, applicability to additional temporal
constraints, and capability in random crossing stress tests, which are
challenging for priority-free timing updates, as coupled passing-order conflicts dominate the problem.

\subsection{Experimental Settings and Benchmark Scenarios}
\label{subsec:exp_settings}

We consider four benchmark classes:
\begin{itemize}
    \item
    \textit{Bottleneck}: a dense queueing scenario through a narrow shared corridor.
    \item
    \textit{Aligned crossing}: closely aligned fixed-routes used to evaluate arrival time coordination in structured traffic.
    \item
    \textit{Graph trajectory}: agents track paths on a directed geometric graph generated from randomly sampled nodes and Delaunay triangulation. Each agent is assigned the shortest path between a start and a goal node. Directed edges are used to avoid opposing direction conflicts along the same segment.
    \item
    \textit{Random crossing}: randomly perturbed waypoint paths through a shared central region, used as a coupled stress test.
\end{itemize}
Fig.~\ref{fig:benchmark_snapshots} shows representative route geometries.

All benchmark scenarios use a workspace with spatial scale approximately $[-10,10]~{\rm m}\times[-10,10]~{\rm m}$. The speed bounds $[v_{\rm min},v_{\rm max}]=[0.02,2.0]~{\rm m/s}$ are used in all cases. Table~\ref{tab:sim_params} lists the default parameters for the proposed method.

\begin{table}[t]
\caption{Default parameter settings}
\label{tab:sim_params}
\centering
\renewcommand{\arraystretch}{1.15}
\begin{tabular}{c c l}
\hline
\textbf{Variable} & \textbf{Value} & \textbf{Description} \\
\hline
$\rho$ & $10^2$ & ADMM penalty parameter \\
$N_{\rm iter}$ & $1000$ & Maximum number of ADMM iterations \\
$M$ & $200$ & Number of temporal grid nodes \\
$\varepsilon_{\rm pri}$ & $10^{-8}$ & Primal residual tolerance \\
$\varepsilon_{\rm safe}$ & $10^{-3}$ & Safety violation tolerance \\
$\epsilon_{\rm wp}$ & $0.1\,{\rm m}$ & Flyby waypoint tracking tolerance \\
$m_{\rm stag}$ & $10$ & Stagnation window size \\
$\varepsilon_{\rm stag}$ & $10^{-2}$ & Stagnation threshold \\
$\alpha_{\rm active}$ & $0.7$ & Inertial gain \\
$\varepsilon_{\rm inc}$ & $1.0$ & Penalty increase tolerance \\
\hline
\end{tabular}
\end{table}

\begin{figure}[!b]
\centering
{\includegraphics[width=1.0\linewidth]{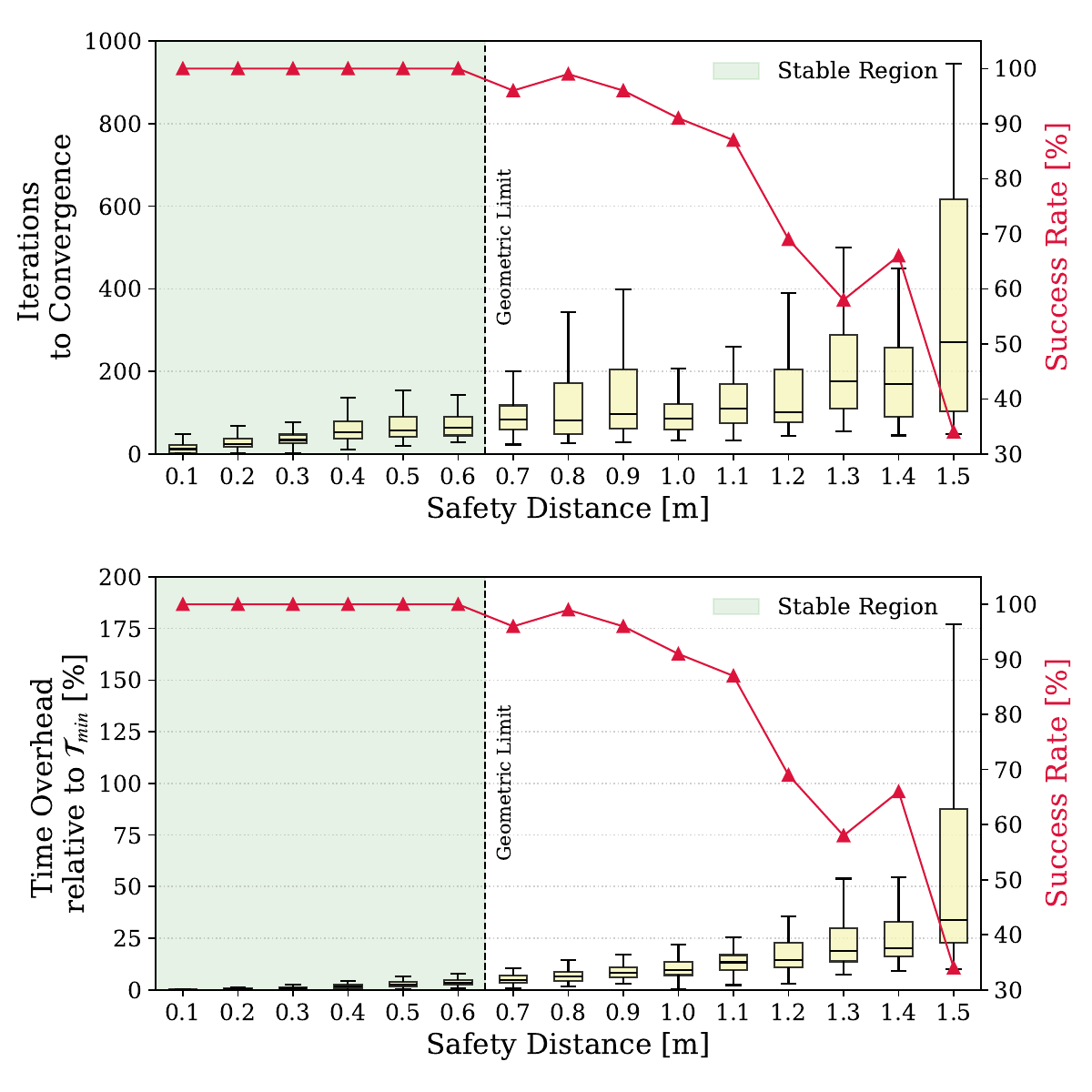}}
\caption{Graph route safety distance sweep for the proposed method. The box plots show the number of iterations and the time overhead, and the red curve shows the success rate. The shaded region indicates safety distances satisfying $d_{\rm safe}<(\sqrt{3}/4)d_{\rm node}$.}
\label{fig:graph_dsafe_sweep}
\end{figure}

The safety distance $d_{\rm safe}$ is selected by scenario. For the bottleneck and random crossing cases, it is determined by the occupancy density $\phi = K\pi(d_{\rm safe}/2)^2/A$, which gives $d_{\rm safe} = 2\sqrt{A\phi/(K\pi)}$, where $A$ is the workspace area. For the aligned-crossing case, we set $\phi=5\%$. For the graph route experiments, $d_{\rm safe} = (\sqrt{3}/4)d_{\rm node}$, where $d_{\rm node}$ is the average distance between network nodes. This value is tied to the geometric scale of the Delaunay network. Fig.~\ref{fig:graph_dsafe_sweep} reports the corresponding safety distance sweep.

We compare the proposed method with three fixed-route baselines:
\textit{mixed-integer programming--second-order cone programming (MIP-SOCP)}, a hierarchical scheduling pipeline that uses mixed-integer
programming for passing-order selection \cite{wu2019temporal};
\textit{Priority planning}, a prioritized planner with
waiting insertion inspired by continuous-time prioritized safe-interval
planning \cite{kasaura2022prioritized}; and
\textit{Fixed-route control barrier function quadratic program (CBF-QP)}, a route-projected CBF-QP that adjusts scalar speeds
along the prescribed route tangents
\cite{mestres2024distributed}.

\subsection{Benchmark Results}
\label{subsec:benchmark_results}

Table~\ref{tab:baseline_standard} compares the proposed method with the three baselines on standard fixed-route benchmarks. Runtime is not reported because the implementations use different prototype solver backends and levels of code optimization. We report success rate, mean arrival time, and maximum arrival time.

\begin{table*}[!t]
\caption{
Baseline comparison on standard fixed-route benchmarks.
Each entry reports the success rate, mean arrival time $\bar{T}$, and
maximum arrival time $T_{\rm max}$
}
\label{tab:baseline_standard}
\centering
\scriptsize
\setlength{\tabcolsep}{2.7pt}
\renewcommand{\arraystretch}{1.20}
\begin{tabular*}{\textwidth}{
@{\extracolsep{\fill}}
l
ccc
!{\vrule width 0.35pt}
ccc
!{\vrule width 0.35pt}
ccc
!{\vrule width 0.35pt}
ccc
@{}
}
\hline
\multirow{2}{*}{\textbf{Scenario}}
&
\multicolumn{3}{c}{\textbf{Proposed}}
&
\multicolumn{3}{c}{\textbf{MIP-SOCP}}
&
\multicolumn{3}{c}{\textbf{Priority planning}}
&
\multicolumn{3}{c}{\textbf{fixed-route CBF}}
\\
\cline{2-13}
&
\textbf{SR}
& $\bar{T}$ [s]
& $T_{\rm max}$ [s]
&
\textbf{SR}
& $\bar{T}$ [s]
& $T_{\rm max}$ [s]
&
\textbf{SR}
& $\bar{T}$ [s]
& $T_{\rm max}$ [s]
&
\textbf{SR}
& $\bar{T}$ [s]
& $T_{\rm max}$ [s]
\\
\hline
Bottleneck, $\phi=1\%$
& $100\%$ & 13.32 & 17.40
& $100\%$ & 15.05 & 20.35
& $100\%$ & \textbf{13.19} & 16.78
& $0\%$   & --    & --
\\
Bottleneck, $\phi=5\%$
& $100\%$ & \textbf{15.33} & 21.53
& $100\%$ & 16.97 & 22.79
& $100\%$ & 15.65 & 21.48
& $0\%$   & --    & --
\\
Bottleneck, $\phi=10\%$
& $100\%$ & \textbf{16.96} & 24.74
& $100\%$ & 18.82 & 27.27
& $100\%$ & 17.07 & 24.58
& $0\%$   & --    & --
\\
\hline
Aligned crossing
& $100\%$ & \textbf{11.33} & 12.72
& $100\%$ & 14.90 & 18.88
& $100\%$ & 11.34 & 13.03
& $0\%$   & --    & --
\\
\hline
Graph trajectory
& $100\%$ & \textbf{9.16} & 11.11
& $97\%$  & 9.54 & 12.16
& $99\%$  & 9.25 & 11.97
& $47\%$  & 9.35 & 11.44
\\
\hline
\end{tabular*}
\end{table*}

\begin{table*}[!t]
\caption{
Comparison on graph route benchmarks with timing constraints. The dwell-time case imposes a uniform dwell duration at every intermediate waypoint. In the fixed-terminal time case, one agent is assigned a target terminal time computed using the nominal speed defined in Section~\ref{subsec:exp_settings}. Each entry reports the success rate, mean arrival time $\bar{T}$, and maximum arrival time $T_{\rm max}$. Arrival time statistics are computed over successful runs
}
\label{tab:timing_constrained_graph}
\centering
\scriptsize
\setlength{\tabcolsep}{2.7pt}
\renewcommand{\arraystretch}{1.20}
\begin{tabular*}{\textwidth}{
@{\extracolsep{\fill}}
l
ccc
!{\vrule width 0.35pt}
ccc
@{}
}
\hline
\multirow{2}{*}{\textbf{Scenario}}
&
\multicolumn{3}{c}{\textbf{Proposed}}
&
\multicolumn{3}{c}{\textbf{Priority planning}}
\\
\cline{2-7}
&
\textbf{SR}
& $\bar{T}$ [s]
& $T_{\rm max}$ [s]
&
\textbf{SR}
& $\bar{T}$ [s]
& $T_{\rm max}$ [s]
\\
\hline
Graph with waypoint dwell time
& $100\%$ & 17.22 & 22.80
& $100\%$ & 17.12 & 24.40
\\
Graph with fixed-terminal time
& $100\%$ & 9.93 & 20.08
& $98\%$ & 10.01 & 20.11
\\
\hline
\end{tabular*}
\end{table*}

In the bottleneck benchmark, priority planning gives a slightly lower mean arrival time at the lowest density, whereas the proposed method gives comparable or lower mean arrival times as the density increases. In the aligned crossing and graph trajectory benchmarks, the proposed method gives the lowest mean arrival time in Table~\ref{tab:baseline_standard} and successfully solves all tested instances. The MIP-SOCP baseline succeeds in most cases but gives larger arrival times in these structured scenarios. The fixed-route CBF baseline, which uses equal priority reactive speed control, fails in most scenarios due to deadlock. We therefore compare the proposed method with the most competitive baseline, the priority planning baseline in graph trajectory scenarios with additional
timing constraints.

Table~\ref{tab:timing_constrained_graph} reports graph trajectory results under additional timing constraints. The comparison is restricted to the proposed method and Priority planning. The MIP-SOCP baseline is omitted because the formulation in \cite{wu2019temporal} does not include these timing constraints.

In the dwell-time case, each agent dwells for $1.0\,\mathrm{s}$ at
every intermediate waypoint. Both methods successfully solve all tested
instances. Priority planning gives a slightly lower mean arrival time, while the proposed method gives a lower maximum arrival time.

In the fixed-terminal time case, we impose a target arrival time on a specific agent. The target time is defined as the duration required to traverse the given route at the mean of the prescribed speed bounds. The proposed method successfully solves all tested instances and yields
slightly lower mean and maximum arrival times than the baseline. In contrast, priority planning fails to find a feasible solution in $2\%$ of the
tested instances.

These results indicate that the proposed priority-free timing
formulation is competitive with the priority planning baseline on
the structured fixed-route benchmarks.

\subsection{Stress Test and Parameter Sensitivity Analysis}
\label{subsec:random_crossing_stress}

\begin{table}[t]
\caption{
Random crossing stress test. SR is computed over all generated instances. Iteration and cost statistics are computed over successful runs and reported as median $[Q_{25},Q_{75}]$
}
\label{tab:random_crossing_stress}
\centering
\scriptsize
\setlength{\tabcolsep}{3.0pt}
\renewcommand{\arraystretch}{1.15}
\begin{tabular*}{\columnwidth}{@{\extracolsep{\fill}} c c c c c @{}}
\hline
\textbf{$K$} &
$\phi$ &
\textbf{SR} &
\textbf{Iter.} &
\textbf{Cost}
\\
\hline
\multirow{3}{*}{3}
  & $1\%$  & $100\%$ & 57 [53, 64]    & 1.21 [1.17, 1.24] \\
  & $5\%$  & $100\%$ & 86 [78, 91]    & 1.56 [1.47, 1.86] \\
  & $10\%$ & $100\%$ & 90 [87, 94]    & 1.69 [1.64, 1.99] \\
\hline
\multirow{3}{*}{7}
  & $1\%$  & $99\%$  & 211 [152, 264] & 1.31 [1.26, 1.41] \\
  & $5\%$  & $100\%$ & 206 [174, 251] & 1.88 [1.70, 2.07] \\
  & $10\%$ & $99\%$  & 273 [234, 346] & 2.49 [2.20, 2.78] \\
\hline
\multirow{3}{*}{11}
  & $1\%$  & $87\%$  & 546 [434, 678] & 1.49 [1.43, 1.56] \\
  & $5\%$  & $64\%$  & 626 [531, 746] & 2.59 [2.27, 2.97] \\
  & $10\%$ & $19\%$  & 730 [656, 830] & 3.72 [3.28, 4.23] \\
\hline
\end{tabular*}
\end{table}

\begin{table}[t]
\caption{
Parameter sensitivity in the random-crossing stress test. SR is
computed over all generated instances, whereas iteration count and cost
are reported over successful runs as the median and interquartile
interval $[Q_{25},Q_{75}]$. Bold entries indicate the default parameter
setting and its corresponding results.
}
\label{tab:param_sensitivity}
\centering
\scriptsize
\setlength{\tabcolsep}{2.5pt}
\renewcommand{\arraystretch}{1.15}
\begin{tabular*}{\columnwidth}{@{\extracolsep{\fill}} l c c c c @{}}
\hline
\textbf{Param.} &
\textbf{Value} &
\textbf{SR} &
\textbf{Iter.} &
\textbf{Cost}
\\
\hline
\multirow{4}{*}{$\alpha_{\rm active}$}
& $0$ & $7\%$  & 847 [728,873] & 1.94 [1.90,1.98] \\
& $0.3$ & $22\%$ & 780 [590,884] & 2.03 [1.96,2.19] \\
& $\mathbf{0.7}$ & $\mathbf{64\%}$ & \textbf{626} [531,746] & \textbf{2.59} [2.27,2.97] \\
& $1.0$ & $96\%$ & 585 [460,775] & 4.54 [3.61,6.34] \\
\hline
\multirow{3}{*}{$\varepsilon_{\rm inc}$}
& $0.5$ & $54\%$ & 652 [569,761] & 2.29 [2.11,2.52] \\
& $\mathbf{1.0}$ & $\mathbf{64\%}$ & \textbf{626} [531,746] & \textbf{2.59} [2.27,2.97] \\
& $2.0$ & $86\%$ & 576 [464,714] & 2.87 [2.56,3.17] \\
\hline
\multirow{3}{*}{$\rho$}
& $10$ & $70\%$ & 594 [496,701] & 2.63 [2.38,3.02] \\
& $\mathbf{100}$ & $\mathbf{64\%}$ & \textbf{626} [531,746] & \textbf{2.59} [2.27,2.97] \\
& $1000$ & $71\%$ & 534 [448,697] & 2.63 [2.26,2.99] \\
\hline
\end{tabular*}
\end{table}

The random crossing benchmark is designed as a stress test for the proposed method. In this scenario, multiple routes overlap in a confined central region, and agents approaching from opposite sides share long path portions. These interactions require coupled passing-order decisions, making the case challenging for a formulation without prescribed priorities. We use this benchmark to assess the limits of the local timing update.

Table~\ref{tab:random_crossing_stress} reports the random crossing stress test. The success rate is computed over all tested instances, while the cost and iteration statistics are conditioned on successful runs. The results show that problem difficulty increases with both the agent count $K$ and the occupancy density $\phi$. For $K=3$, all instances at the tested densities are solved within the iteration budget. For $K=7$, the success rate remains high, with larger iteration counts and costs. For $K=11$, the success rate decreases from $87\%$ at $\phi=1\%$ to
$19\%$ at $\phi=10\%$. A failed run means that the residual and safety criteria were not satisfied within the iteration budget. These trends indicate that, as
congestion increases, the proposed priority-free approach has more difficulty entering a feasible passing-order basin within the prescribed iteration budget.

The collision avoidance auxiliary variable update includes an inertial safeguard to mitigate such local stagnation. Table~\ref{tab:param_sensitivity} evaluates its effect on the $K=11$, $\phi=5\%$ random crossing set. The results show a tradeoff between success rate and cost. With $\alpha_{\rm active}=0$, the success rate is $7\%$, and the low cost reflects the easier instances that were solved. Increasing $\alpha_{\rm active}$ improves the success rate, up to $96\%$ at $\alpha_{\rm active}=1.0$, but also increases the returned cost. A similar trend appears for $\varepsilon_{\rm inc}$: a looser reset condition improves the success rate but yields higher cost schedules. The ADMM penalty $\rho$ has a smaller effect over the tested range.

The random crossing results indicate that highly symmetric, densely coupled crossing patterns are challenging for the proposed priority-free local timing update. In these cases, many agents interact near the same region, and the iteration can become trapped in a local passing-order basin. The non-inertial update alone is often insufficient to escape such stagnation. The inertial safeguard improves the success rate by continuing a useful correction direction when progress stalls, as shown in Table~\ref{tab:param_sensitivity}. However, this benefit comes with a cost trade-off, and some dense cases still remain unsolved within the iteration budget. Addressing these passing-order bottlenecks without explicit priority
assignment is left for future work.

\section{Conclusion}

This paper presents a priority-free temporal optimization framework for fixed-route multi-agent systems. The proposed framework optimizes waypoint passage times to generate velocity profiles along given routes, subject to speed limits, pairwise safety constraints, and temporal requirements such as waypoint dwell times and fixed terminal times.

In the standard benchmarks, the proposed method yielded feasible schedules across all tested instances, whereas order-based baselines occasionally failed to compute feasible schedules in the graph trajectory scenarios. Furthermore, the proposed method demonstrated slightly better performance in most cases and achieved comparable results in environments with dwell time constraints and fixed-terminal times. These results demonstrate that the proposed priority-free framework is highly competitive with conventional order-based approaches.

The results also clarify the intended limitations of our approach. Dense random crossing is inherently order-dominated because of its structural symmetry, making it difficult to establish feasible yielding patterns without explicit sequence selection. In this stress test, the proposed local timing update solved all tested \(K=3\) cases, but for \(K=11\) the success rate decreased from \(87\%\) at \(\phi=1\%\) to \(19\%\) at \(\phi=10\%\). The inertial safeguard mitigated some of this stagnation: increasing \(\alpha_{\rm active}\) from \(0.7\) to \(1.0\) increased the success rate from \(64\%\) to \(96\%\), but the median cost increased from \(2.59\) to \(4.54\). These observations clarify a limitation of the proposed priority-free formulation. While the method is effective as a temporal optimizer for fixed-routes with waypoint-level timing constraints, dense order-dominated cases may require explicit mechanisms for resolving local passing-order stagnation. Future work will address these bottlenecks associated with resolving passing orders without explicit priority variables while retaining the waypoint time formulation and its support for dwell time and terminal time constraints.

\section*{Acknowledgment}
During the preparation of this work, the authors used ChatGPT-5.5 to improve grammar, spelling, and phrasing, without changing technical contents. After using this tool, the authors reviewed and edited the content as needed and take full responsibility for the content of the published article.

\bibliographystyle{IEEEtran}
\bibliography{ref}

\end{document}